# Risk assessment and observation of driver with pedestrian using instantaneous heart rate and HRV

**Riku Kikuta[1,2], Daniel Carruth[1,2], Reuben Burch[1,2], John Ball[1,3], and Ichiro Kageyama[4]**

[1] Center for Advanced Vehicular Systems, Mississippi State University, Starkville, MS 39759, USA

[2] Bagley College of Engineering, Industrial and System Engineering, Mississippi State University, Starkville, MS 39759, USA

[3] Bagley College of Engineering, Electrical and Computer Engineering, Mississippi State University, Starkville, MS 39759, USA

[4] College of Industrial Technology, Nihon University, 1-2-1, Izumi-cho, Narashino, Chiba, Japan

**ABSTRACT**

Currently, human drivers outperform self-driving vehicles in many conditions such as collision avoidance. Therefore, understanding human driver behaviour in these conditions will provide insight for future autonomous vehicles. For understanding driver behaviour, risk assessment is applied so far as one of the approaches by using both subjective and objective measurement. Subjective measurement methods such as questionnaires may provide insight into driver risk assessment but there is often significant variability between drivers. Physiological measurements such as heart rate (HR), electroencephalogram (EEG), and electromyogram (EMG) provide more objective measurements of driver risk assessment. HR is often used for measuring driver's risk assessment based on observed correlations between HR and risk perception. Previous work has used HR to measure driver's risk assessment in self-driving systems, but pedestrian dynamics is not considered for the research. In this study, we observed driver's behaviour in certain scenarios which have pedestrian on driving simulator. The scenarios have safe/unsafe situations (i.e., pedestrian crosses road and vehicle may hit pedestrian in one scenario), HR analysis in time/frequency domain is processed for risk assessment. As a result, HR analysis in frequency domain shows certain reasonability for driver risk assessment when driver has pedestrian in its traffic.

## INTRODUCTION

In 2021, there were more than 40,000 traffic fatalities in the US and 90% of the accidents were caused by human error according to the National Highway Traffic Safety Administration (NHTSA, 2022). Advanced driver-assistance systems (ADAS) and self-driving systems have been under active development for many years to reduce such traffic accidents. While these systems are intended to provide support to drivers, a driver can sometimes outperform the systems in certain situations requiring complex decisions in a moment. Understanding real human driving behavior is necessary to develop comfortable self-driving systems (Miyajima, 2016). Therefore, methods that allow developers to better understand driver behavior should lead to ways to reduce traffic accidents. Driver models, mathematical models that replicate the functionality of the driver during driving, have been used to understand and analyze human driving behavior (Kageyama, 2018; Li, 2013; Wang, 2008). Understanding driver risk assessment is often used to build driver models. In many cases, driver risk assessment is evaluated through subjective measurement using tools such as questionnaires. However, it is difficult to develop a general understanding of driver risk assessment as it varies depending on each driver's individual perception. Therefore, some research uses objective measurements such as driver's heart rate as a proxy for measuring risk assessment and building an accurate driver model. A driver's instantaneous heart rate is correlated with their perception of risk, and perception of risk leads the driver to use steering or braking to minimize their current risk (Kageyama, 2018; Raksincharoensak, 2016). The previous research shows that a risk perception driver model can describe driver behavior in limited conditions such as the Double Lane Change scenario and collision avoidance. However, most research only tests one condition, so it is difficult to assess the differences between responses in safe and unsafe conditions. Hence, this study investigates driver's heart rate in safe, unsafe, and dynamically changing scenarios in a driving simulator. For instance, we have a scenario in which risk of collision disappears when the pedestrian suddenly stops at the edge of the crosswalk. If the pedestrian does not stop, there will be a collision. The driver is expected to perceive significant risk from the pedestrian before the pedestrian stops, so they should apply the brake. However, the driver perceives less or no risk when the pedestrian stops at the curb and will stop braking.

      Most research which uses heart rate to measure risk assessment analyzes heart rate in time-domain. However, it is expected that heart rate in time-domain analysis is not appropriate for driving scenarios as the heart rate may not fluctuate. Frequency domain analysis is one of the ways to understand human's psychological state such as stress, nervousness, and risk (Novani, 2018; Pham, 2021). In this paper both time-domain and frequency-domain analysis of heart rate is used to understand how the drivers respond to the pedestrian in safe/unsafe traffic situations. This study aims to observe drivers' behavior and evaluate the difference between each scenario according to differences in observed heart rate. The driver's heart rate is collected using a Polar H 10 heart rate sensor. Nine participants were recruited for data collection. The data collection is approved by Mississippi State University Institutional Review Board (protocol ID: IRB-22-296).

## DRIVING SIMULATOR

The driving simulator is constructed using Epic Games Unreal Engine. Figure 1 shows the layout of the road environment and Figure 2 shows the road environment as it appears in the simulator. The environment consists of a two-lane road, one crosswalk, one pedestrian, and a sidewalk on each side. The simulator is run on a Windows 10 computer and uses a steering wheel and accelerator and brake pedals. Figure 3 shows an overview of the driving simulator and the heart rate sensor (Polar H 10). In the simulator, participants are asked to maintain 40 mph (approximately 60 km/h, 18 m/s) and stay in their current lane. When the driver sees the pedestrian, the driver tries to avoid a collision with the pedestrian by pressing the brake pedal. The driving simulator records these outputs for analysis: velocity, acceleration/deceleration, and position.

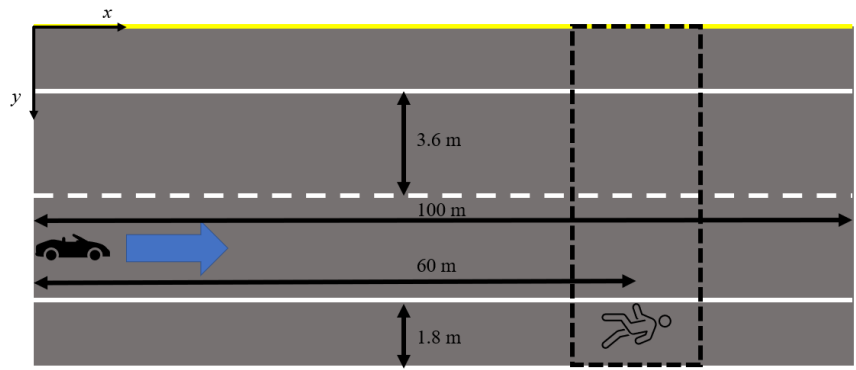

**Figure 1:** Diagram of the road environment.

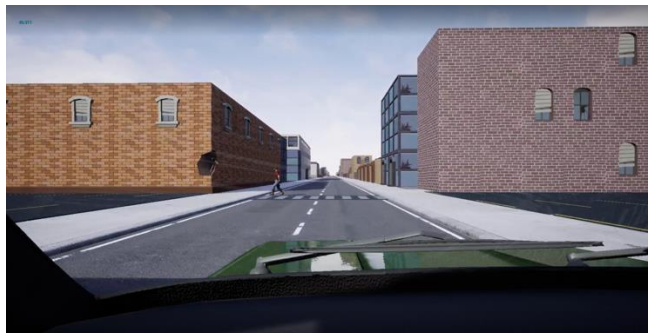

**Figure 2:** Road environment as it appears in the driving simulator.

## Simulation Scenarios

In this study, we focus on the response of the driver and its differences according to safe/unsafe conditions which are determined by pedestrian dynamics. Nine driving simulator scenarios were defined including safe and unsafe scenarios as well as differences in pedestrian dynamics. The pedestrian appears 60 m from the vehicle (where simulation time = around 240 second), and its dynamics varies in each scenario.

**Scenario 1:** The pedestrian crosses the crosswalk and the vehicle will hit the pedestrian unless the driver brakes hard (Unsafe crossing).

**Scenario 2:** The pedestrian crosses the crosswalk before the vehicle reaches the crosswalk. When the driver recognises the pedestrian, the pedestrian is in the left lane (Safe crossing 1).

**Scenario 3:** The pedestrian crosses the crosswalk after the vehicle passes the crosswalk. The pedestrian is far away when the driver recognises the pedestrian (Safe crossing 2). Basically, the pedestrian cannot reach the crosswalk before the vehicle reaches it.

**Scenario 4:** The pedestrian is far away when the driver recognises the pedestrian, and the pedestrian will stop at the edge of the sidewalk (Safe stop).

**Scenario 5:** The pedestrian starts stopped at the edge of the sidewalk (Stopping). Even if the pedestrian is stopped, they might start to cross at any time, so there is still some risk of collision for the driver.

**Scenario 6:** The pedestrian starts in the center of the right road lane and does not move (Unsafe stopping 1). Therefore, there will be a collision unless the driver brakes hard. This is similar to scenario 1, but the pedestrian is not moving in this scenario. This scenario aims to observe the effect of different pedestrian dynamics in an unsafe scenario.

**Scenario 7:** The pedestrian crosses the crosswalk but stops at the center of the right lane suddenly (Unsafe stopping 2). If the pedestrian did not stop, there would not be a collision. This scenario aims to observe how drivers respond to sudden change increasing risk of a collision.

**Scenario 8:** The pedestrian moves to cross the crosswalk but stops at the edge of the sidewalk (Safe stopping). If the pedestrian does not stop, there would be a collision (same as scenario 1). This scenario observes how driver response when the risk of collision suddenly disappears.

**Scenario 9:** The pedestrian crosses the lane ahead of the vehicle, but turns back, and returns to the lane (Unsafe return). In this scenario, it initially seems there will be no collision, but the risk of collision suddenly appears. This scenario is like scenario 2 and 7 but with differences in pedestrian dynamics.

## PARTICIPANTS

Nine drivers participated in this experiment with each participant performing three of the scenarios. The participants are recruited according to these criteria: Have a valid US driver license, age between 18-65, fluent in English, and not at risk for an epileptic seizure. Each participant was assigned randomly to one of the three groups where each group is assigned three simulation scenarios (i.e., group A has scenario 1, 4, and 7) including at least one unsafe scenario. Table 1 shows the list of participants and scenarios belonging to each group. Hence, this experiment has three data for each scenario.

**Table 1.** List of participants and scenarios belonging to each group

| Group   | Scenario         | Participants |
|---------|------------------|--------------|
| Group A | Scenario 1, 4, 7 | A, F, H      |
| Group B | Scenario 2, 5, 6 | B, C, G      |
| Group C | Scenario 3, 8, 9 | D, E, I      |

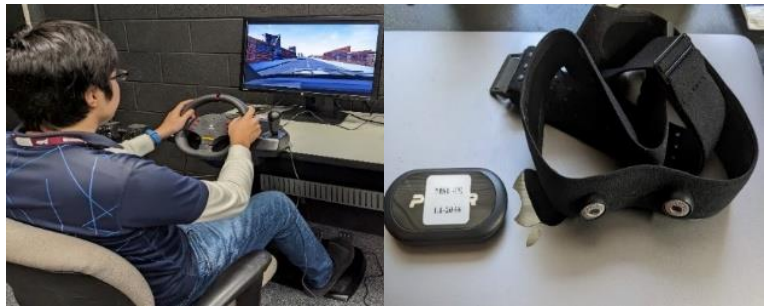

**Figure 3:** Overview of the driving simulator and heart rate sensor Polar H10.

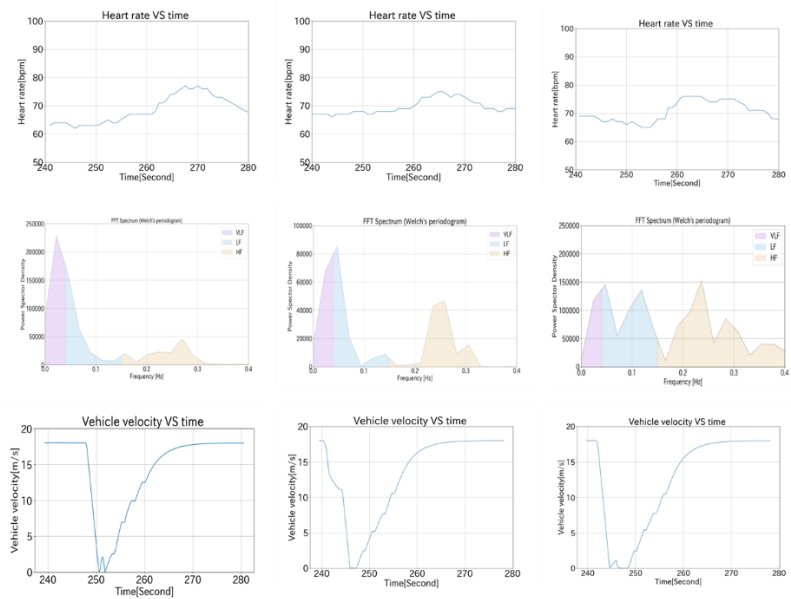

**Figure 4:** Example of results of participant A (left: Scenario 1, center: Scenario 4, right: Scenario 7).

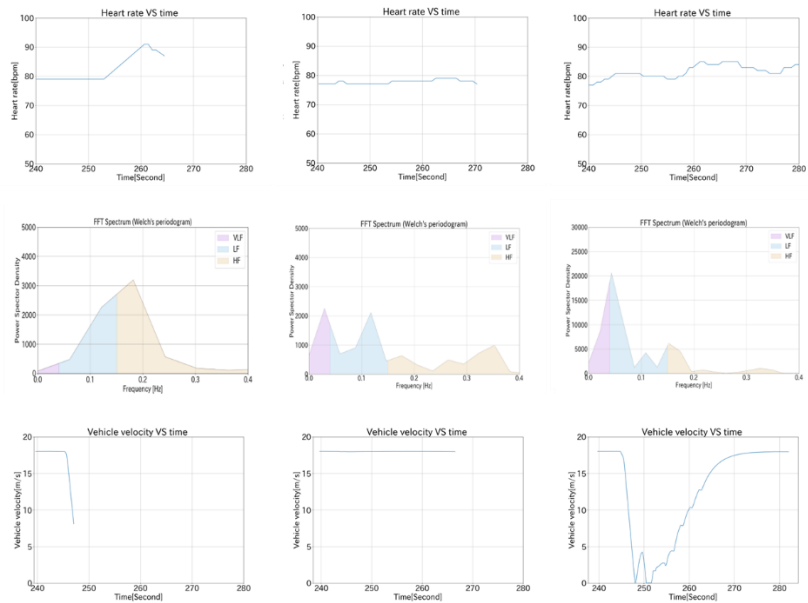

**Figure 5:** Example of results of participant F (left: Scenario 1, center: Scenario 4, right: Scenario 7).

## RESULTS

### Experimental Results in Time Domain

Figure 4 and Figure 5 show representative results for two participants that completed scenarios 1, 4, and 7 in the study. Figure 4 shows results of participant A and Figure 5 shows results of participant F. Participant A brakes in all scenarios and their heart rate increases in that time even though the situation is safe. Participant A perceives risk in scenario 4 despite the safe situation. Their heart rate increases, and they brake to avoid a collision even though the pedestrian will not reach the crossroad before the vehicle passes.

Participant F is in position to hit the pedestrian in scenario 1 and therefore, the vehicle velocity in scenario 1 ends at approximately 247 seconds as the driving simulator is automatically terminated when the driver would hit the pedestrian. As Figure 5 shows, participant F does not brake when the situation is safe (scenario 4), and their heart rate is steady indicating that the participant does not perceive risk from the pedestrian. Furthermore, participant F brakes in scenario 7 and their heart rate increases like participant A. Therefore, the participant perceives risk in the unsafe scenarios (scenario 1 and 7) and does not perceive risk in the safe scenario (scenario 4). As a result, we confirmed that for some participants heart rate increases in time domain and vehicle response (velocity) match in both safe and unsafe scenarios.

## Data processing for Frequency Domain

It is expected that analysis of heart rate for risk assessment only in time domain is difficult as heart rate changes are delayed from vehicle response and sometimes do not show clear changes. Therefore, it is important to discuss the results in frequency domain for risk assessment. Fast Fourier Transform (FFT) has been used for frequency domain analysis not only in heart rate, but also vibration engineering, signal processing, and crashworthiness as it can extract time-independent features. HRV is used for frequency domain analysis and has been associated with human stress, nervousness, and risk (Novani, 2018). HRV is a measure of the variation in time between each heartbeat and it is sometimes called the R-R interval. For instance, if a current HRV is 600, it means the heart beats once a 0.6 second. Also, HRV is collected not at a certain interval as it is calculated every beats. It is difficult to process FFT in HRV data as HRV is not collected at a certain interval, and FFT requires evenly sampled data. Therefore, this study interpolates the HRV data by 3-order spline interpolation. After the interpolation, Power Spector Density (PSD) is calculated by Welch periodogram and FFT.

The graphs in the center column of Figures 4 and 5 show the HRV results in the frequency domain. The purple region shows VLF (very low frequency band power): 0.0-0.04 Hz, the blue region shows LF (low frequency band power):0.04-0.15 Hz, and the yellow region shows HF (high frequency band power): 0.15-0.4 Hz. Those three regions are usually used for human psychological assessment such as stress, nervousness, and risk. For instance, if there is peak value in both LF and HF, this indicates that the subject was stressed. If there is peak value only in LF or VLF, this indicates that the subject was not stressed. In this study, baseline HRV is collected by Polar H10 heart rate monitor before driving any scenarios. The baseline HRV is compared with each scenario's HRV to determine if there is a difference between them, especially regarding peaks. Hence, this study uses the three regions to confirm if there are differences in perception of risk across the scenarios.

**Experimental Results in Frequency Domain**

In this study, we obtained HRV PSD results by FFT. The results of participant A show a clear difference between safe and unsafe situations. There is a peak value in HF region in scenario 1 and 7 which indicates that participant A perceives risk from the pedestrian. Although there is a peak value in scenario 4, the value is quite small compared to the other two regions indicating that the participant does not perceive risk from the pedestrian. However, despite the small peak in the region, participant A brakes in scenario 4 indicating a disconnect between the frequency domain analysis and the vehicle dynamics. Similarly, the results of participant F in frequency domain and vehicle dynamics do not match. In scenario 1, the frequency domain analysis indicates a peak value in HF region and the participant brakes. However, in scenario 4, there is a peak value in the HF region, but the participant does not brake.

**DISCUSSION**

This study analyzed heart rate data both in time domain and frequency domain to determine if there is clear difference between different traffic situations that may relate to driver dynamics. The results in time domain indicate that the driver's heart rate increases when the driver brakes. In other words, risk perception and vehicle response match when the heart rate is evaluated in time domain. In contrast, the results of frequency domain analysis do not match vehicle dynamics even though there is a difference in HRV for safe and unsafe scenarios. Frequency domain analysis may provide a physiological measure of risk perception of driver, it cannot be used alone to predict driving behavior (vehicle dynamics).

It is expected that each driver has own their risk threshold (i.e., a risk taker has a high risk threshold which means that they tend to accept higher risk situations even though they can perceive the risk, the risk averse have a low risk threshold which means that they tend to reduce the current risk even though there is small risk), and frequency domain results may be affected by the threshold effect. Participant F does not brake in scenario 4 as it is safe, despite the frequency domain analysis indicating that the participant does perceive some small risk from the pedestrian. Therefore, participant F might accept the risk even though the risk is perceived by participant as this behaviour may not affect against other traffics and its flow.

**CONCLUSION**

In this study, we examined the relationship between driver responses and heart rate in both time and frequency domains during safe and unsafe interactions with pedestrians. In time domain analysis, there are clear difference between heart rate and vehicle dynamics, and therefore, it is expected that heart rate analysis in time domain is reasonable for measuring risk assessment of the driver and understanding driver's response. Although frequency domain does not show clear relationship to vehicle dynamics possibly due to driver's risk threshold or specific parameters which affect driver's response, differences in HRV were observed between safe and unsafe conditions and therefore, HRV in frequency domain analysis still provides insight into driver risk assessment.

Since this research indicates us that each driver's specific parameters like threshold may affect against vehicle dynamics, future observation and evaluation of expert driver HR, HRV and driving behaviour should inform development of self-driving systems as it is expected to show better performance. Therefore, a driver model based on expert drivers would be expected to result in better performance which causes less traffic accidents and more comfort.

**REFERENCES**

U.S. Department of Transportation National Highway Traffic Safety Administration. (May 5 2022) Early Estimates of Motor Vehicle Traffic Fatalities And Fatality Rate by Sub-Categories in 2021,
https://crashstats.nhtsa.dot.gov/Api/Public/ViewPublication/813298.
C. Miyajima, K. Takeda. (2016) Driver-Behavior Modeling Using On-Road Driving Data: A new application for behavior signal processing, IEEE Signal Processing Magazine, vol. 33, no. 6, pp. 14-21.
I. Kageyama, Y. Kurigayagawa, A. Tsubouchi. (2018) Study on construction of driver model for obstacle avoidance using risk potential, *IAVST: International Symposium on Dynamics of Vehicles on Roads and Tracks,* vol. 1.
L. Bi. (2013) A Driver Lateral and Longitudinal Control Model Based on Queuing Network Cognitive Architecture, *Fourth Global Congress on Intelligent Systems,* pp. 274-278.
J. Wang, L. Zhang, D. Zhang, K. Li. (2013) An Adaptive Longitudinal Driving Assistance System Based on Driver Characteristics, IEEE Transactions on Intelligent Transportation Systems, vol. 14, no. 1, pp. 1-12.
P. Raksincharoensak, T. Hasegawa, M, Nagai. (2016) Motion planning and control of autonomous driving intelligence system based on risk potential optimization framework, *International Journal of Automotive Engineering,* vol. 7, pp. 53-60.
N. Novani, L.Arief, R. Anjasmara, A. Prihatmanto. (2018) Heart Rate Variability Frequency Domain for Detection of Metal Stress Using Support Vector Machine, *2018 International Conference on Information Technology Systems and Innovation (ICITSI),* vol. 7, pp. 520-526.
T. Pham, Z. Lau,S. Chen, D. Makowski. (2021) Heart Rate Variability in Psychology: A Review of HRV Indices and an Analysis Tutorial, *Sensors 2021,* vol. 21, no.12.